\documentclass[sigconf]{acmart}
\usepackage{epsfig}
\usepackage{graphicx}
\usepackage{amsmath}
\usepackage{amssymb}
\usepackage{threeparttable}

\usepackage{array}
\usepackage{subfigure}
\usepackage{caption}
\usepackage{epstopdf}
\usepackage{hyperref}
\usepackage{enumitem}

\usepackage{balance}

\def\BibTeX{{\rm B\kern-.05em{\sc i\kern-.025em b}\kern-.08emT\kern-.1667em\lower.7ex\hbox{E}\kern-.125emX}}

\copyrightyear{2019}
\acmYear{2019}
\acmConference[MM '19]{Proceedings of the 27th ACM International Conference on Multimedia}{October 21--25, 2019}{Nice, France}
\acmBooktitle{Proceedings of the 27th ACM International Conference on Multimedia (MM '19), October 21--25, 2019, Nice, France}
\acmPrice{15.00}
\acmDOI{10.1145/3343031.3350985}
\acmISBN{978-1-4503-6889-6/19/10}

\begin{document}

\fancyhead{}

\title{Sentence Specified Dynamic Video Thumbnail Generation}

\author{Yitian Yuan}
\authornote{This work was done while Yitian Yuan was a Research Intern at Tencent AI Lab.}
\email{yyt18@mails.tsinghua.edu.cn}
\affiliation{
  \institution{Tsinghua-Berkeley Shenzhen Institute, Tsinghua University}
}

\author{Lin Ma}
\email{forest.linma@gmail.com}
\affiliation{
  \institution{Tencent AI Lab}
}

\author{Wenwu Zhu}
\authornote{Corresponding author.}
\email{wwzhu@tsinghua.edu.cn}
\affiliation{
  \institution{Department of Computer Science and Technology \& Tsinghua-Berkeley Shenzhen Institute, Tsinghua University}
}

%
\renewcommand{\shortauthors}{Yuan, et al.}

%
\begin{abstract}
With the tremendous growth of videos over the Internet, video thumbnails, providing video content previews, are becoming increasingly crucial to influencing users' online searching experiences. Conventional video thumbnails are generated once purely based on the visual characteristics of videos, and then displayed as requested. Hence, such video thumbnails, without considering the users' searching intentions, cannot provide a meaningful snapshot of the video contents that users concern. In this paper, we define a distinctively new task, namely sentence specified dynamic video thumbnail generation, where the generated thumbnails not only provide a concise preview of the original video contents but also dynamically relate to the users' searching intentions with semantic correspondences to the users' query sentences. To tackle such a challenging task, we propose a novel graph convolved video thumbnail pointer (GTP). Specifically, GTP leverages a sentence specified video graph convolutional network to model both the sentence-video semantic interaction and the internal video relationships incorporated with the sentence information, based on which a temporal conditioned pointer network is then introduced to sequentially generate the sentence specified video thumbnails. Moreover, we annotate a new dataset based on ActivityNet Captions for the proposed new task, which consists of 10,000+ video-sentence pairs with each accompanied by an annotated sentence specified video thumbnail. We demonstrate that our proposed GTP outperforms several baseline methods on the created dataset, and thus believe that our initial results along with the release of the new dataset will inspire further research on sentence specified dynamic video thumbnail generation. Dataset and code are available at \textbf{\texttt{\url{https://github.com/yytzsy/GTP}}}.
\end{abstract}



%

%
\maketitle

\section{Introduction}

\begin{figure}
\centering
\setlength{\abovecaptionskip}{0.cm}
\setlength{\belowcaptionskip}{-0.2cm}
\includegraphics[width=0.95\columnwidth]{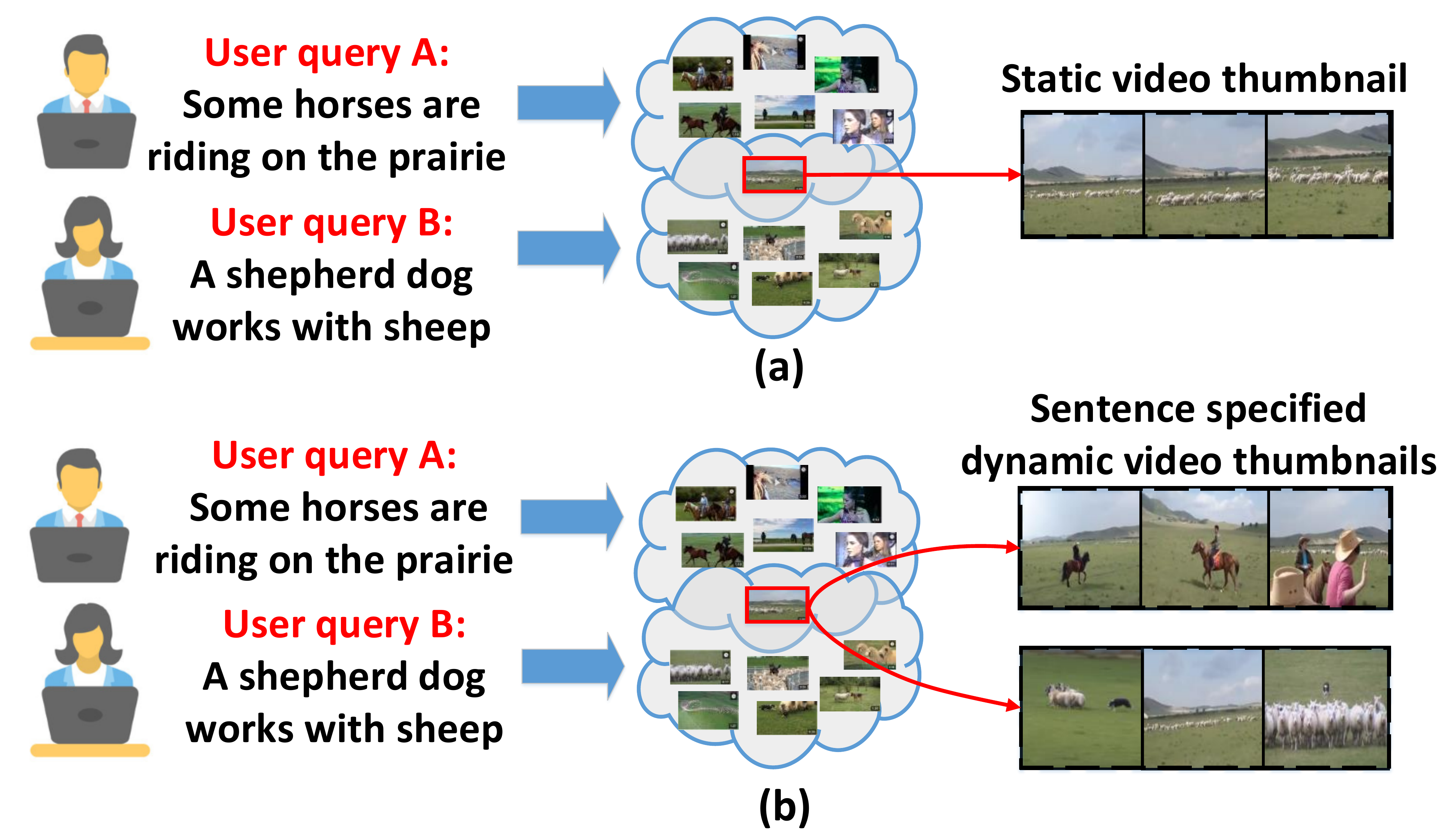}
\caption{The comparison between traditional static video thumbnail and our proposed sentence specified dynamic video thumbnails for online video searching scenarios.}
\label{fig:introduction}
\end{figure}

Tremendous popularity of video websites and social networks has stimulated a massive growth of videos over the Internet. In face of this data deluge, video thumbnail \cite{Liu2015Multi,Song2016To}, as a commonly used technology to provide viewers a condensed and straightforward preview about the video contents, is becoming increasingly crucial to influencing users' online searching and browsing experiences. Traditionally, one single key frame is extracted from an original video as its thumbnail, which only conveys limited information and cannot provide a vivid preview of the video. Therefore, some popular video websites, like YouTube \footnote{https://www.youtube.com/}, start to trim a short segment from a video as the video thumbnail, which provides a snapshot of what the video is about.

From picking one single key frame to trimming one segment, video thumbnails are becoming more expressive. However, there are still some problems that have been overlooked before. Currently, most video thumbnails are yielded purely based on their visual characteristics (\textit{e.g.} visual quality, representativeness), while regardless of the users' search intentions \cite{Dufaux2004Keyframe,Kang2005To,Luo2009Towards,Hasebe2010Video,Song2016To,wang2011extracting}. For example, user A and user B in Figure~\ref{fig:introduction}(a) search online videos based on two different queries ``\texttt{\small Some horses are riding on the prairie}'' and ``\texttt{\small A shepherd dog works with sheep}''. It can be observed that there is one video existing in both returned video pools. However, the pre-determined video thumbnail, even in the form of a video segment, only presents the scene of sheep, which partially relates to the query of user B and is irrelevant to the search intention of user A. We regard such a video thumbnail to be ``static'' to the users' queries. By browsing such video thumbnails, users still cannot decide whether the video contains the meaningful and desired information they need, which will greatly influence the efficiency and experience of online video searching.

Nowadays, a thread of works  \cite{Liu2011Query,Liu2015Multi,Vasudevan2017Query} take users' queries into consideration for generating video thumbnails. On the one hand, such methods limit video thumbnails in the form of a single key frame without considering video temporal characteristics, thus making the generated video thumbnails less expressive. On the other hand, users' queries employed in these methods are often confined to single words or short phrases, which cannot accommodate general and flexible users' searching intentions in the form of natural language sentences. Besides the above, another thread of works \cite{Hendricks2017Localizing,Gao2017TALL,liu2018attentive,chen2018temporally} which aim to trim a single consecutive video segment from a video according to the given natural language query, can also apply to the video thumbnail generation task. However, such methods mainly focus on modeling video-sentence semantic correlation while ignore global video contents and internal video relationships, making the trimmed segment not comprehensive enough as a video thumbnail to express the video contents.

Based on the above considerations, in this paper, we define a distinctively new task, namely sentence specified dynamic video thumbnail generation. First, a video is evenly split into a sequence of short video clips. Afterward, we exploit the semantic relationships between these video clips as well as their matching behaviors with the query sentence, and finally select and concatenate several video clips to compose the final video thumbnail. Different from the traditional video thumbnails which are pre-determined offline, as shown in Figure \ref{fig:introduction}(b), our video thumbnails are dynamically generated concerning different sentence queries.

The sentence specified dynamic video thumbnail generation is a very challenging task. Firstly, natural sentence query and video are different kinds of sequential data with rich semantic meanings. Therefore, their matching relationships are quite complicated and need to be modeled in a fine-grained manner, so as to generate video thumbnails that conform to users' search intentions. Secondly, as a video thumbnail can be composed by several video clips, how to model the internal semantic relationships within videos and make the selected video clips semantically coherent with the overall video contents is worthy of further considerations.

To address the aforementioned challenges, we propose a novel graph convolved video thumbnail pointer (GTP), which can generate a semantically meaningful and coherent video thumbnail from an input video and meanwhile make the yielded thumbnail semantically relevant to the natural sentence query.
Specifically, GTP first establishes a word-by-clip attention interaction between the sentence query and video sequence, and then performs a fine-grained semantic coupling of these two modalities. Afterward, based on the yielded sentence-video interaction features, a graph convolutional network (GCN) \cite{Kipf2016Semi} is performed to model the sentence specified relationships between different video clips, and further supports the in-video reasoning under the sentence semantics. Finally, a novel temporal conditioned pointer network, which takes the graph convolved features as input, is proposed to sequentially generate the video thumbnail and meanwhile preserve its semantic coherence.

Another major obstacle for sentence specified dynamic video thumbnail generation is the lack of dataset which contains pairs of video and sentence descriptions, as well as the associated sentence specified video thumbnails. To this end, we create a new dataset by annotating thumbnails for videos in the ActivityNet Captions~\cite{caba2015activitynet,Krishna2017Dense} dataset. We take one video segment in ActivityNet Captions and its associated caption as our required video and sentence pair, and annotate the video thumbnail for the video segment, making the thumbnail semantically relevant to the caption. In total, our dataset consists of 10,000+ video-sentence pairs collected from about 4,000 videos and their captions in the ActivityNet Captions dataset.

In summary, our contributions are four-folds:
\vspace{-2mm}
\begin{itemize}[leftmargin=*]
\item We introduce a novel task, namely sentence specified dynamic video thumbnail generation, aiming at dynamically selecting and concatenating video clips from an original video to generate one video thumbnail, which not only provides a concise preview of the original video but also semantically corresponds to the given sentence description.
\item We propose a novel graph convolved video thumbnail pointer (GTP) to tackle the sentence specified dynamic video thumbnail generation problem. A sentence specified video graph convolutional network is designed to exploit the complicated semantic relationships within the sentence and video sequence, based on which a temporal conditioned pointer network is proposed to sequentially generate the video thumbnail and meanwhile preserve its semantic coherence.
\item We annotate video thumbnails for videos in the ActivityNet Captions dataset, and create a new dataset to facilitate the research on sentence specified dynamic video thumbnail generation.
\item We validate the effectiveness of our proposed GTP model on the newly created dataset and achieve superior performance against the competing methods.
\end{itemize}

\section{Related Work}
\noindent \textbf{Text Independent Video Thumbnail Generation}.  Most conventional video thumbnail generation methods \cite{gao2009thematic,mei2009video,Song2016To,Dufaux2004Keyframe,Kang2005To,Hasebe2010Video} have focused on learning the characteristics of video thumbnails purely from visual contents, regardless of the user input textual queries. Particularly, Gao \textit{et al.} \cite{gao2009thematic} proposed a thematic video thumbnail selection algorithm, which constructs a visual theme model to capture the visual commodities shared between video key frames and an extra set of web images searched by the keywords from the video. Key frames with the highest similarities to the visual theme can be selected as the final video thumbnails. Song \textit{et al.} \cite{Song2016To} presented an automatic thumbnail selection system which selects attractive thumbnails by analyzing various objective and subjective metrics (\textit{e.g.,} visual quality and aesthetics) of video frames. They performed clustering analysis to determine the relevance between the video thumbnail and video content, and further investigated that the selection of a good thumbnail highly relies on objective visual quality metrics, such as frame texture and sharpness.

Recently, Song \textit{et al.} \cite{Gygli2016Video2GIF} further introduced the problem of automatically generating animated gifs from videos. Gifs are short looping video segments of no sound and can present the expressive video contents to users, and therefore can be regarded as a new form of video thumbnails. To solve the gif generation problem, they proposed a robust deep RankNet, which models video content popularity and quality and further generates a ranking list of video segments according to their suitabilities as a gif. While the above methods can select visually qualified key frames or segments from videos to represent video contents, they ignore the user intentions for searching videos, which may not be adequate to satisfy the users' online searching and browsing experiences.

\noindent \textbf{Text Specified Video Thumbnail Generation}. Recently, some researchers start to investigate how to generate video thumbnails according to textual user queries \cite{Liu2011Query,Liu2015Multi,Vasudevan2017Query}. Huang \textit{et al.} \cite{Liu2011Query} proposed a query-specific thumbnail selection algorithm that extracts a frame being both representative of the video contents and specific to the intent of the user's query. The matching relations between query words and frame contents are captured by a shallow dual cross-media relevance model \cite{liu2007dual} adapted from the image annotation problem. Liu \textit{et al.} \cite{Liu2015Multi} employed a deep visual-semantic embedding model (VSEM) to measure the relevance between the query and video frames by embedding them into a latent semantic space. Hence, key frames in the video are ranked by their distances to the given query in the learned latent space, and the top-ranked frames are selected as the final video thumbnail. Based on VSEM, Vasudevan \textit{et al.} \cite{Vasudevan2017Query} further proposed a quality-aware relevance estimation model (QARE) which can capture the query-independent frame-quality properties in the visual semantic embedding procedure. The frame-quality properties are characterized separately by one dimension in the common latent semantic space. Thus, their video thumbnail selection is done by using both the query dependent relevance scores and query-independent quality scores of video frames.

 Most of the above text specified video thumbnail generation methods are largely based on the multi-modal semantic matching framework \cite{Frome2013DeViSE,pan2014click}, which is originally designed for image searching or tagging. Due to the lack of datasets customized for video thumbnail generation, these methods can only leverage other image annotation datasets such as Clickture \cite{Hua2013Clickage} to train their models. With such image-based framework and dataset, a lot of important video specific characteristics such as video temporal relationships are not fully explored and leveraged, which inevitably hurts the effectiveness of the video thumbnail generation. Moreover, the user queries are often confined to single words or phrases, which also cannot accommodate the general and flexible user sentence queries.

 \noindent \textbf{Temporal Sentence Localization in Video.} Given an untrimmed video and a natural language sentence query, temporal sentence localization in video aims to identify the start and end points of one video segment, which semantically matches the given sentence query \cite{Hendricks2017Localizing,Gao2017TALL,liu2018attentive,chen2018temporally,yuan2019find,chen2019localizing,chen2019weaklysupervised}. To solve this problem, Hendricks \textit{et al.} firstly presented a Moment Context Network (MCN) \cite{Hendricks2017Localizing} to match video segments with sentence query in a multi-modal latent space, where the temporal endpoint features of video segments are also incorporated to enhance the localization performance. Gao \textit{et al.} proposed a Cross-Modal Temporal Regression Localizer (CTRL) \cite{Gao2017TALL}, which extended the object detection methodologies \cite{girshick14CVPR,Girshick2015Fast} in spatial dimensions to temporal dimension. They firstly sampled several candidate video segments from video and fused the sentence information with each of these segments. Then based on the fused multimodal features, the temporal boundaries of these segments were adjusted to the target positions with a localization regression network. Liu \textit{et al.} proposed a Attentive Cross-Modal Retrieval Network (ACRN) \cite{liu2018attentive}. The ACRN enhanced the CTRL architecture with a memory attention mechanism, in which the visual information mentioned in the query was emphasized and further incorporated to the context of each candidate segment.

Our proposed sentence specified dynamic video thumbnail generation task is different from the temporal sentence localization task. For temporal sentence localization, it is assumed that the given sentence query only corresponds to one single video segment, which consists of one or several consecutive video clips. However, for dynamic video thumbnail generation, the predicted thumbnails can be composed of several temporally inconsecutive but semantically coherent video clips. More importantly, the temporal sentence localization task mainly emphasizes on modeling the semantic correlation between video and sentence. While for sentence specified video thumbnail generation, the generated video thumbnail not only should have close relationships with the sentence query, but also needs to provide a straightforward preview of the overall video contents. Therefore, the global video information, such as the semantic relationships between different video clips, needs to be considered for generating the dynamic video thumbnail.

\section{Proposed Approach}

Given a video $V$ and a sentence $S$, the task of sentence specified dynamic video thumbnail generation aims to select a set of video clips $\left\{ v_i \right\}$ from $V$, which are semantically relevant to the sentence $S$ and will be concatenated together as the final video thumbnail. Each video is first represented as $\mathbf{V} = \left\{ \mathbf{v}_t \right\}_{t=1}^{T}$, where $\mathbf{v}_t$ denotes the representation of the $t$-th video clip, and $T$ is the total number of clips. Accordingly, each sentence is represented as $\mathbf{S} = \left\{ \mathbf{w}_n \right\}_{n=1}^{N}$, where $\mathbf{w}_n$ is the embedding of the $n$-th word in the sentence and $N$ denotes the total number of words.

We propose a novel graph convolved video thumbnail pointer (GTP), to tackle the sentence specified dynamic video thumbnail generation problem. As illustrated in Figure~\ref{fig:framework}, GTP, which takes the video and sentence features $\mathbf{V}$ and $\mathbf{S}$ as inputs, consists of three modules: (1) video and sentence encoders, (2) sentence specified video graph convolutional network and (3) temporal conditioned pointer network. Please note that the three modules are closely coordinated and can thus be trained in an end-to-end fashion.

\vspace{-1mm}
\subsection{Video and Sentence Encoders}
Considering the sequential characteristics of the video and sentence representations, two bi-directional gated recurrent units (BiGRUs)~\cite{Cho2014Learning} are used to encode these two modalities, respectively:
\begin{equation}
\setlength{\abovedisplayskip}{0pt}
\setlength{\belowdisplayskip}{0pt}
\small{
\begin{split}
    &  \mathbf{u}_t^V = \text{BiGRU}_V(\mathbf{u}_{t-1}^V,\mathbf{u}_{t+1}^V,\mathbf{v}_t), \\
    &  \mathbf{u}_n^S = \text{BiGRU}_S(\mathbf{u}_{n-1}^S,\mathbf{u}_{n+1}^S,\mathbf{w}_n).
\end{split}
}
\end{equation}
Due to the behaviors of BiGRU, the output hidden states, namely $\mathbf{U}^V = [\mathbf{u}_1^V, \cdots, \mathbf{u}_T^V]$ and $\mathbf{U}^S = [ \mathbf{u}_1^S, \cdots, \mathbf{u}_N^S ]$,  encode and aggregate the flexible contexts of the video and sentence, respectively.

\subsection{Sentence Specified Video Graph Convolutional Network}
\begin{figure}
\setlength{\abovecaptionskip}{0.cm}
\setlength{\belowcaptionskip}{-0.1cm}
\centering
\includegraphics[width=1.0\columnwidth]{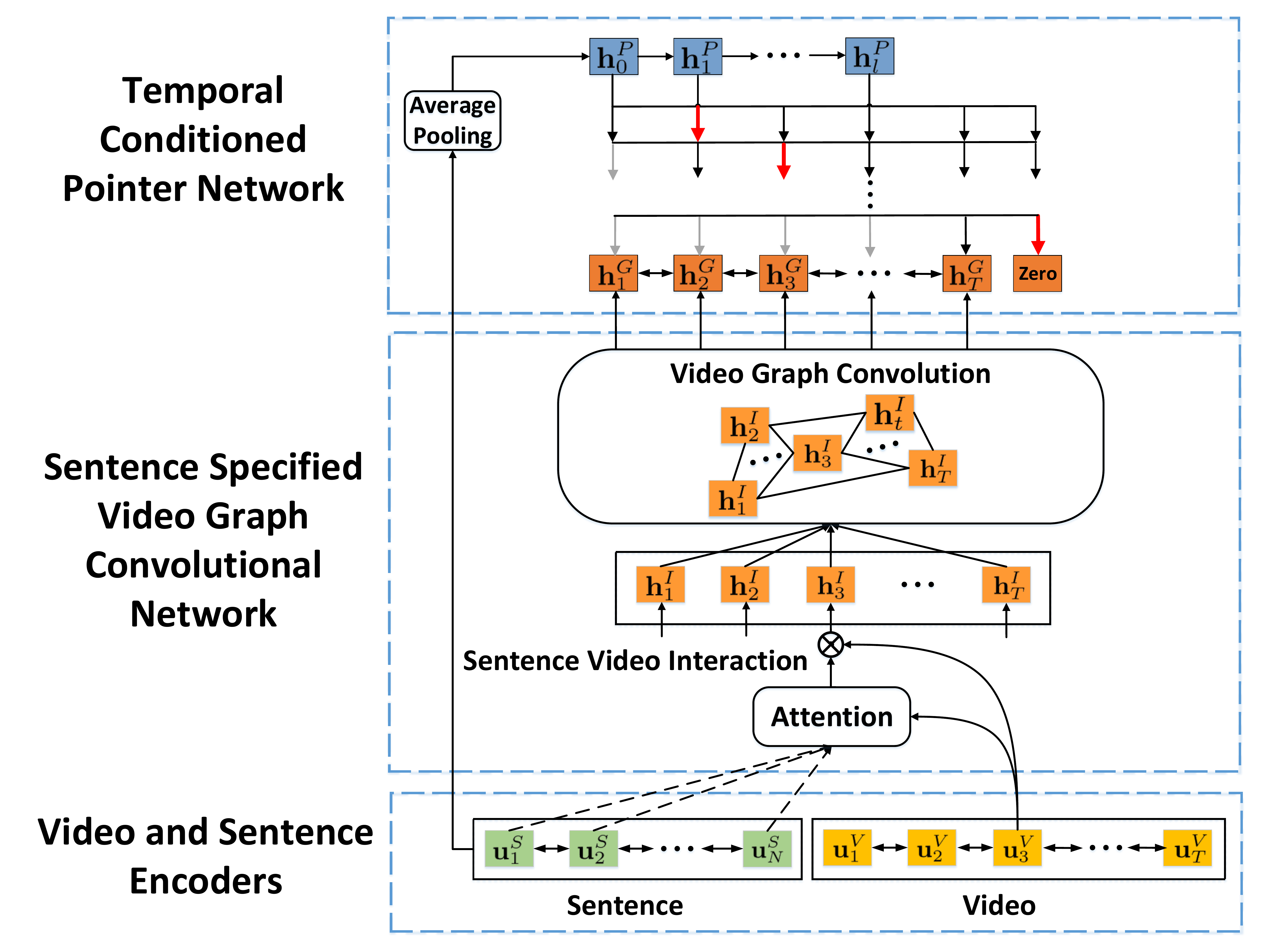}
\caption{The architecture of our GTP model, which consists of three modules. First, the video and sentence encoders aggregate the contextual evidences from the video clip representations and word embeddings of the sentence query, respectively. Second, the sentence specified video graph convolutional network establishes the fine-grained word-by-clip interaction between the sentence and video, and leverages a GCN to further exploit the sentence specified video clip relationships. Finally, the temporal conditioned pointer network predicts and concatenates  the video clips to yield the video thumbnail in a sequential manner.}
\label{fig:framework}
\end{figure}

Relying on the encoded video $\mathbf{U}^V$ and sentence $\mathbf{U}^S$ representations, as shown in the middle part of Figure \ref{fig:framework}, the sentence video interaction and the video graph convolution modules are stacked together to exploit the fine-grained sentence video semantic relationships and the sentence specified video clip relationships, respectively.

\noindent \textbf{Sentence Video Interaction.}
To fully exploit the fine-grained interaction between sentence and video, we propose to attentively summarize and incorporate the sentence information regarding each video clip. Specifically, the soft attention mechanism \cite{softAttention} is used to generate the attention weights $\left\{a_n^t\right\}_{n=1}^N$ of one video clip with respect to all the words in the sentence:
\begin{equation} \small
\begin{split}
     \beta_n^t = \mathbf{w}^T {\rm tanh} \left(  \mathbf{W}^I_s  \mathbf{u}_n^S + \mathbf{W}^I_v \mathbf{u}_t^V + \mathbf{b}_a^I \right), \ \ \ \ a_n^t = \frac{{\rm exp}(\beta_n^t)}{ \sum_{n=1}^{N} {\rm exp}(\beta_n^t)},
\end{split}
\label{eq:interaction_1}
\end{equation}
where $\mathbf{w}^T$, $\mathbf{W}^I_s$, $\mathbf{W}^I_v$, and $\mathbf{b}_a^I$ are the learnable parameters. The clip-specific sentence representation $\mathbf{c}_t^S$ is subsequently computed by aggregating the word features with the yielded attention weights:
\begin{equation} \small
\begin{split}
    \mathbf{c}_t^S & = \sum_{n=1}^N a_n^t \mathbf{u}_n^S.
\end{split}
\label{eq:interaction_1_2}
\end{equation}
Finally, we concatenate each video clip feature with its clip-specific sentence feature, and feed the concatenated vector to a fully-connected (FC) layer:
\begin{equation} \small
 \mathbf{h}^{I}_t = \sigma \left(\mathbf{W}^I_f \big( \mathbf{u}_t^V \| \mathbf{c}_t^S \big) + \mathbf{b}^I_f \right),
\label{eq:interaction_2}
\end{equation}
where $\sigma$ is the nonlinear activation function, and $\mathbf{W}^I_f$ and $\mathbf{b}^I_f$ are the parameters of the FC layer. The yielded $\mathbf{H}^I = [\mathbf{h}_1^I, \cdots, \mathbf{h}_T^I]$, denoted as the sentence-video interaction features, dynamically encodes the fine-grained word-by-clip matching relationships between the sentence and video.

\noindent \textbf{Video Graph Convolution.}
In our sentence specified dynamic video thumbnail generation task, the generated video thumbnails should not only have close relationships with the sentence semantics, but also need to provide a content preview of the overall video. Therefore, with the sentence-video interaction features, we further model the sentence specified relationships between different video clips by a graph convolutional network \cite{Kipf2016Semi}, so as to take the global video contents into consideration when generating video thumbnails. Specifically, we represent the video as a graph structure, where each node $\mathbf{h}_t^I$ in the graph represents one video clip incorporated with sentence information, and the edge between each pair of nodes represents their sentence specified semantic similarity or affinity  $\mathbf{F}_{ij} = {\mathbf{h}_i^I}^T \mathbf{h}_j^I$. After computing the affinity matrix $\mathbf{F}$, we perform normalization on each row of the matrix to ensure that the sum of the edge values connected to one node be 1~\cite{Vaswani2017Attention,Wang2018Videos}:
\begin{equation} \small
    \mathbf{G}_{ij} = \frac{{\rm exp} (\lambda \mathbf{F}_{ij}) } {\sum_{j=1}^T {\rm exp} (\lambda \mathbf{F}_{ij}) },
\label{GCN_equation}
\end{equation}
where $\lambda$ is the scaling factor. $\mathbf{G} \in \mathbb{R}^{T \times T}$ is regarded as the adjacency matrix representing the constructed sentence specified video clip graph.

Based on the adjacency matrix $\mathbf{G}$, the graph convolution operation is performed, which computes the response of a node based on its neighbors defined by the above sentence specified graph relationships:
\begin{equation} \small
\mathbf{Z} = (\mathbf{G}+\mathbf{I}) \mathbf{X} \mathbf{W}^G,
\setlength{\abovedisplayskip}{0pt}
\setlength{\belowdisplayskip}{3pt}
\end{equation}
where $\mathbf{I} \in \mathbb{R}^{T \times T}$ is the identity matrix to emphasize the self-interaction of each node. $\mathbf{X} \in \mathbb{R}^{T \times d}$ is the representations of all the graph nodes. $\mathbf{W}^G \in \mathbb{R}^{d \times d}$ is the learnable weight matrix for performing the  convolution operation. The output $\mathbf{Z}$ is of the same dimension as the input $\mathbf{X}$. As such, the graph convolution operation can be stacked into multiple layers. After each layer of graph convolution, the Layer Normalization~\cite{ba2016layer} and nonlinear activation are performed before $\mathbf{Z}$ is forwarded to the next layer. Thus, the graph convolution process can be regarded as performing information passing inside our built graph, or as linking the relevant video clips under the sentence semantics.

In our video graph convolution, the input of the first layer of convolution is the sentence-video interaction features, \textit{i.e.,} $\mathbf{X} = \mathbf{H}^I$, and the output of the last layer of convolution is defined as the graph convolved video features $\mathbf{H}^G = [\mathbf{h}_1^G, \cdots, \mathbf{h}_T^G]$.

\vspace{-1mm}
\subsection{Temporal Conditioned Pointer Network}

\begin{figure}
\setlength{\abovecaptionskip}{0.cm}
\setlength{\belowcaptionskip}{-0.cm}
\centering
\includegraphics[width=3.3in]{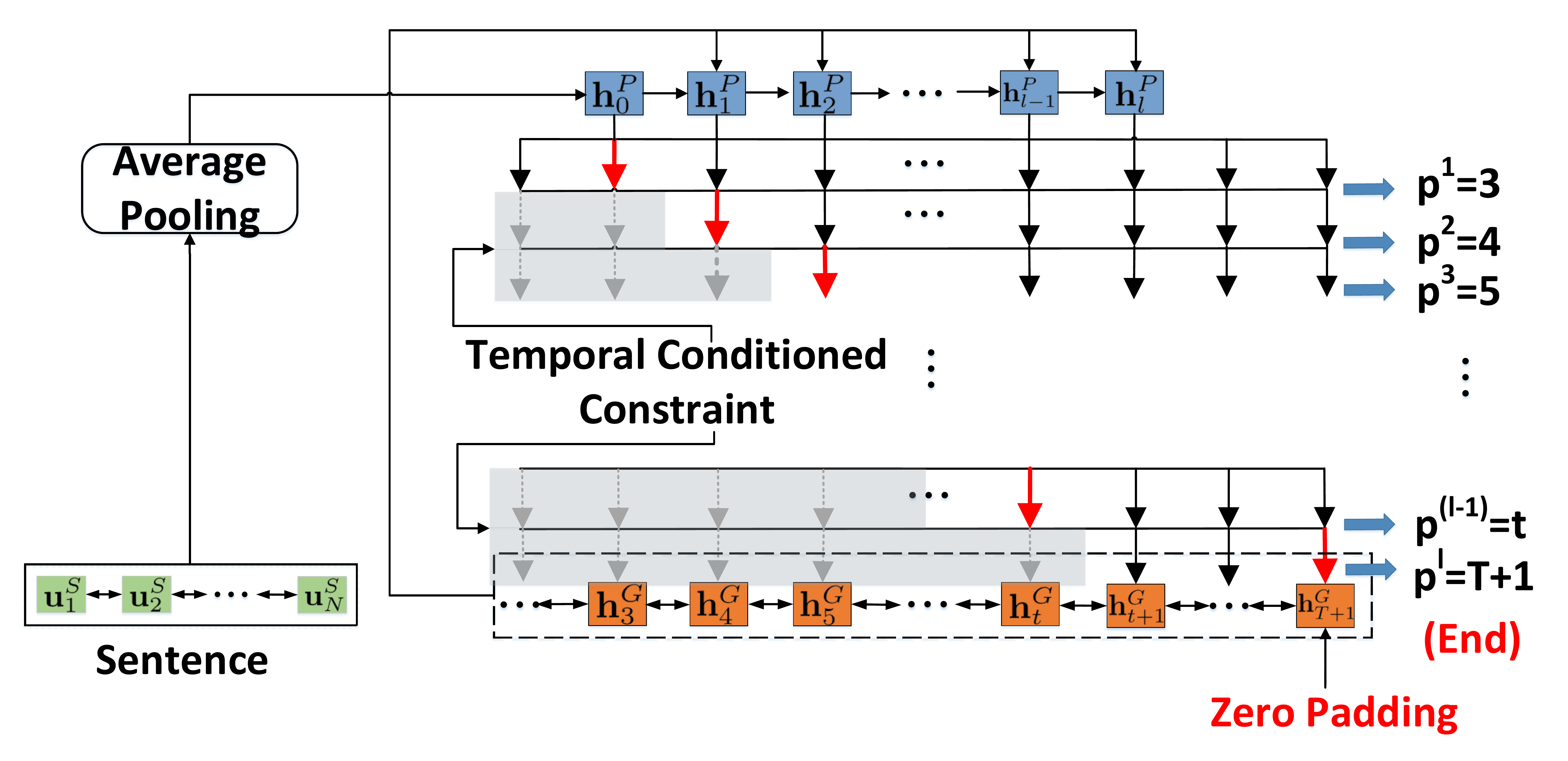}
\caption{ The detailed architecture of the proposed temporal conditioned pointer network. From top to bottom, each red arrow points out the selected video clip in the sequential video thumbnail generation procedure. The video clip selection stops until it points to the zero padding state $\mathbf{h}^G_{T+1}$ at a certain time-step. Under the temporal conditioned constraint, the gray bar in each row indicates the video clips that will not be selected at each time-step.}
\label{fig:pointer_detail}
\vspace{-3mm}
\end{figure}

Based on the graph convolved video features, we design a novel temporal conditioned pointer network shown in Figure~\ref{fig:pointer_detail}, which sequentially outputs a list of integers $\mathbf{p} = (p^1,\cdots,p^j,\cdots)$ indicating the selected video clips to be concatenated as the desired video thumbnail.

Specifically, another BiGRU is used to aggregate the graph convolved video features as $\Tilde{\mathbf{H}}^G = [\mathbf{H}^G;\mathbf{h}_{T+1}^G]$, where $\mathbf{h}_{T+1}^G = \mathbf{0}$ is a padding token used to indicate the end of the sequential video clip selection. To determine $p^j$, a temporal conditioned attention mechanism is proposed to compute an attention vector $\mathbf{e}^j \in \mathbb{R}^{T+1}$, where $e^j_t$ indicates the probability of selecting the $t$-th video clip as the $j$-th clip to compose the final video thumbnail:
\begin{equation}
\small{
\begin{split}
    & s_t^j = \mathbf{w}^T {\rm tanh}   \left(  \mathbf{W}^P_g \mathbf{h}_t^{G} + \mathbf{W}^P_h \mathbf{h}_{j-1}^P + \mathbf{b}^P \right), \\
     & e_t^j = \frac{ m_t^j {\rm exp} (s_t^j)}{\sum_{t=1}^{T+1}  m_t^j {\rm exp} (s_t^j)} \  {\rm with}  \
     m_t^j =
   \begin{cases}
   0 &\mbox{if $t \leq p^{j-1}$} \\
   1 &\mbox{if $t > p^{j-1}$} \\
   \end{cases}, \\
     &p^j = \ {\rm argmax}(e_1^j, \cdots, e_{T+1}^j),
\end{split}
\label{eq:constraint}
\setlength{\abovedisplayskip}{0pt}
\setlength{\belowdisplayskip}{0pt}
}
\end{equation}
where $\mathbf{h}_{j-1}^P$ is the hidden state of the temporal conditioned pointer network, which is realized by a GRU:
\begin{equation}
\small{
 \mathbf{h}_{j}^P = {\rm GRU} (\mathbf{h}_{j-1}^P, \sum_{t=1}^{T+1} e^j_t \mathbf{h}^G_t).
}
\setlength{\abovedisplayskip}{0pt}
\setlength{\belowdisplayskip}{0pt}
\end{equation}
At each time-step, the input is yielded by attentively summarizing $\mathbf{H}^G$ regarding the generated probabilities $\mathbf{e}^j$.  $\mathbf{h}_{0}^P$ is initialized by the average pooling of the sentence representation.

Compared with the general pointer network~\cite{Vinyals2015Pointer}, as denoted in Eq (\ref{eq:constraint}), a temporal conditioned constraint, fulfilled via a binary attention mask $m^j_t$, is applied on $s_t^j$ when generating the corresponding attention weight $e_t^j$. In this way, if the position of the previously selected video clip is $p^{j-1}$, the video clips before $p^{j-1}$ will not be considered and deactivated by setting $m^j_t$ to 0 (as illustrated in the gray region of Figure~\ref{fig:pointer_detail}). On the contrary, the general pointer network will choose an already selected clip again or a video clip before the already selected clips. The disordered chosen video clips will break the logical relationships in the video and inevitably hurt the performance of the pointer network in the following time-steps. The proposed temporal conditioned constraint naturally solves the problem by introducing the attention mask, which ensures the generated thumbnail to be temporally consistent with the original video, therefore providing users a semantically coherent preview of the video contents. Moreover, it is worth noting that our proposed temporal conditioned pointer network makes the video clip selection quite flexible, and even inconsecutive video clips can be grouped together to compose the final video thumbnail. Besides, the lengths of the thumbnails are also no need to be limited to a fixed value.

\vspace{-1mm}
\subsection{Training and Inference}
The training samples collected in $\Gamma = \left\{ (V, S, \mathbf{B})\right\}$ for sentence specified dynamic video thumbnail generation are video-sentence-annotation triples. Specifically, each video $V$ is associated with a sentence annotation $(S,\mathbf{B})$, where $S$ is the sentence description used for video thumbnail generation, and $\mathbf{B} \in \mathbb{R}^{T \times K}$ is a ground-truth annotation matrix with binary entries. $T$ is the number of video clips in $V$ and $K$ is the maximal number of video clips that can be contained in a video thumbnail. $\mathbf{B}_t^k$ is set to 1 when the $t$-th video clip in video $V$ is selected as the $k$-th video clip in the video thumbnail. Otherwise, $\mathbf{B}_t^k$ is set to 0.

For a training sample $(V,S,\mathbf{B})$ in $\Gamma$, the objective for video thumbnail generation is given by $L(V,S,\mathbf{B})$:
\begin{equation}
\small{
    L(V,S,\mathbf{B}) = - \sum_{k=1}^K \sum_{t=1}^T \mathbf{B}_t^k {\rm log}  (e_t^k).
}
\end{equation}
Here $e_t^k$ is the predicted selection probability of the $t$-th video clip at the $k$-th step in our proposed temporal conditioned pointer network, as denoted in Section 3.3.

In training, the objective $L$ will back-propagate to all the fully-coupled three modules of GTP. For all the training samples in $\Gamma$, the objective is defined as:
\begin{equation}
\small{
    L_{\Gamma} = \sum_{(V,S,\mathbf{B}) \in \Gamma} L(V,S,\mathbf{B}).
}
\end{equation}
During the inference stage, we first pre-process the input video and sentence description to acquire the video clip and word embedding features, then feed the features into our proposed graph convolved video thumbnail pointer, and finally obtain the predicted positions of the selected video clips. These clips are sequentially concatenated together and constitute the dynamic video thumbnail.

\section{Sentence Specified Video Thumbnail Dataset}
A major challenge for sentence specified dynamic video thumbnail generation is that there is a lack of large-scale dataset which consists of video and sentence pairs, as well as the corresponding sentence-related video thumbnail. To mitigate this issue, we annotate a new dataset based on the ActivityNet Captions \cite{Krishna2017Dense} dataset for our proposed new task.

Each video in ActivityNet Captions is annotated by several sentence captions, with each caption summarizing the content of a specific video segment with explicit starting and ending points in the video. We randomly choose 4,000 videos from ActivityNet Captions, and then trim the video segment for each caption from these chosen videos. The trimmed video segments of less than 20-second length are dropped, and the rest segments with their corresponding captions are collected to form our required video-sentence pairs. We further ask several participants to annotate the video thumbnails for these collected videos. For the convenience of annotation, we set up a website to annotate the video thumbnails. When annotating, participants will watch the video-sentence pair simultaneously. They are required to read the sentence and watch the video first, and then select no more than 5 clips from the video to constitute the final video thumbnail. To speed up the annotation, we split the original video into clips of 2-second length and place these clips on the website in the chronological order. The participants only need to click the clips to indicate their selections.

Through the aforementioned data collection and annotation procedures, we finally acquire 10,204 video-sentence pairs in total, and ensure that each pair is accompanied by 4 video thumbnail annotations from different participants. We randomly choose 70$\%$ of the collected video-sentence pairs for training, 15$\%$ for validation, and the remaining 15$\%$ for testing. Since there are 4 video thumbnail annotations for each video-sentence pair, we take the annotated video thumbnail with the highest consistency among the 4 annotations as the ground-truth during the training stage. While in the testing stage, the predicted video thumbnail will be evaluated with respect to all the 4 annotations. For more details and analysis of our created dataset, please refer to the supplemental material \footnote{https://github.com/yytzsy/GTP/blob/master/ACM\_MM19\_Supplemental\_Material.pdf}.

\vspace{-1mm}
\section{Experiments}
In this section, we begin by describing baseline methods and experimental settings, followed by the experimental results on the sentence specified dynamic video thumbnail generation task.
\subsection{Baseline Methods}
We compare our proposed GTP model against the following state-of-the-art video thumbnail generation methods, specifically BeautThumb \cite{Song2016To}, RankNet \cite{Gygli2016Video2GIF}, VSEM \cite{Liu2015Multi}, and QARE \cite{Vasudevan2017Query}. BeautThumb and RankNet are text independent models which generate video thumbnails by purely relying on visual characteristics of video frames. We directly run the source codes\footnote{Code for BeautThumb: https://github.com/yahoo/hecate; Code for RankNet: https://github.com/gyglim/video2gif\_code}, and concatenate the top-5 ranked video clips as the video thumbnail. VSEM and QARE are text specified models, which learn a joint embedding of video clips and query sentences, and thereby select video thumbnails according to their distances with the sentences. Since both VSEM and QARE only focus on selecting key frames from videos as the thumbnails, we adapt the selection unit of these two methods from video frame to video clip, and the top-5 ranked video clips are concatenated together as the final video thumbnail.

In addition, we also apply two temporal sentence localization methods CTRL \cite{Gao2017TALL} and ACRN \cite{liu2018attentive} to the proposed sentence specified dynamic video thumbnail generation task, and evaluate their results on our created dataset. In the setting of temporal sentence localization in video, one sentence query only refers to one single video segment. However, the annotated video thumbnail in our created dataset may be composed of several inconsecutive video clips. In order to generate corresponding ground truth for temporal sentence localization in our created dataset, for each sentence query, we merge each group of continuous annotated video clips into a video segment, and take the longest video segment as the ground truth for temporal sentence localization.

\subsection{Experimental Settings}

\textbf{Evaluation Metrics.} We assess the quality of a generated video thumbnail by measuring the agreement between the video clips within it and the video clips within the ground-truth annotations. Specifically, for the $k$-th video-sentence sample in the testing set, we denote $A_i^k$ as the set of selected video clips in the $i$-th ground-truth video thumbnail, and $P^k$ as the set of video clips within the generated video thumbnail. The precision, recall, and IoU scores between $A_i^k$ and $P^k$ are computed as
${Precision}_i^k = \frac{ \| Intersection(P^k,A_i^k) \|}{\| P^k\|}$, ${Recall}_i^k = \frac{ \| Intersection(P^k,A_i^k) \|}{\| A_i^k\|}$, ${IoU}_i^k = \frac{ \| Intersection(P^k,A_i^k)\|}{ \| Union( P^k, A_i^k) \|}$. Finally, the overall video thumbnail generation results are evaluated by the average Precision, Recall, F1 and IoU scores among all the M testing samples, as follows:
\begin{equation} \small
Precision = \frac{1}{M} \sum_{k=1}^{M} \max_{i \in \left\{ 1,2,3,4 \right\}} {Precision}_i^k ,
\end{equation}
\begin{equation} \small
Recall  = \frac{1}{M} \sum_{k=1}^{M} \max_{i \in \left\{ 1,2,3,4 \right\}} {Recall}_i^k,
\end{equation}
\begin{equation} \small
F1 = \frac{1}{M}    \sum_{k=1}^{M} \max_{i \in \left\{ 1,2,3,4 \right\}} \frac{2 \times {Precision}_i^k \times {Recall}_i^k}{{Precision}_i^k + {Recall}_i^k} ,
\end{equation}
\begin{equation} \small
IoU =  \frac{1}{M} \sum_{k=1}^{M} \max_{i \in \left\{ 1,2,3,4 \right\}}  {IoU}_i^k.
\end{equation}

\noindent \textbf{Implementation Details.} We evenly split each video into 2-second video clips, and encode each clip with the released C3D \cite{tran2015learning} features by ActivityNet Challenge 2016\footnote{http://activity-net.org/challenges/2016/download.html}.
For sentences, we tokenize each sentence by Standford CoreNLP~\cite{Manning2014The}, and use  Glove~\cite{Pennington2014Glove}  to initialize the word embedding with dimension as 300. The words not found in Glove are randomly initialized. The hidden state dimensions of all GRUs are set as 256. As for the video graph convolution, we set the number of the graph convolution layer as 2, and the scaling factor $\lambda$ as 150. The initial learning rate is set to 0.001, and is gradually decayed over time.

\subsection{Performance Comparisons}

\begin{table}[!tb]\small
\setlength{\abovecaptionskip}{0.cm}
\setlength{\belowcaptionskip}{-0.1cm}
\centering
\caption{ \small Performance comparisons of different methods on our created dataset.}
\begin{tabular}{m{1.9cm} m{0.9cm}<{\centering} m{0.9cm}<{\centering} m{0.9cm}<{\centering} m{0.9cm}<{\centering}}
\hline
Method & Precision & Recall  & F1  & IoU \\
\hline
Random &  0.3409 & 0.3971 & 0.3604 & 0.2379\\
BeautThumb~\cite{Song2016To} & 0.3639 & 0.4217 & 0.3837 & 0.2544 \\
RankNet~\cite{Gygli2016Video2GIF}  &  0.3790 &  0.4443 & 0.4013 & 0.2770 \\
VSEM~\cite{Liu2015Multi} & 0.4142 & 0.4849 & 0.4386 & 0.3098 \\
QARE~\cite{Vasudevan2017Query} &  0.4050 & 0.4744 & 0.4285 & 0.2986 \\
CTRL~\cite{Gao2017TALL} & 0.4933 & 0.4124 & 0.4303 & 0.3084 \\
ACRN~\cite{liu2018attentive} & 0.4967 & 0.4328  & 0.4456 & 0.3271 \\
\hline
\bfseries{GTP}  & \textbf{0.5055} & \textbf{0.5742} & \textbf{0.5285} & \textbf{0.3933} \\
\hline
\end{tabular}
\label{table:Performance_Comparison}
\end{table}

 Table \ref{table:Performance_Comparison} illustrates the video thumbnail generation results of different methods on our created dataset. First, with randomly selecting 5 video clips to constitute the thumbnail, the Random setting performs the worst. Other methods, including our proposed GTP can indeed learn to produce meaningful video thumbnails. Second, the text specified methods VSEM, QARE and GTP achieve much better results than the text independent ones BeautThumb and RankNet. It verifies that incorporating sentence information is beneficial to choose the semantic meaningful video thumbnails for the sentence specified video thumbnail generation task. Third, among the three text specified video thumbnail generation methods, our GTP performs substantially better than VSEM and QARE. Compared with separately matching sentence and each video clip in VSEM and QARE, our GTP establishes a deeper semantic coupling between sentence and video, and captures the sentence specified video clip relations with graph convolution. Moreover, the temporal conditioned pointer network can further preserve the temporal ordering and semantic coherence of the selected video clips. As such, the generated video thumbnail by our proposed GTP is not only semantic related to the sentence description, but also coherent with the overall video contents, and thus demonstrates a significant better performance.

Moreover, as illustrated in Table \ref{table:Performance_Comparison}, the two temporal sentence localization methods, namely  CTRL and ACRN, achieve inferior results compared to our proposed GTP model. Both ACRN and CTRL mainly focus on modeling semantic correlations between videos and sentence queries, while neglect global video contents and internal video relationships, and can only localize one single segment from one video. Even though the predicted video segment may have close relationships to the given sentence query and make relatively high precision value, the single video segment may not be representative enough to cover other meaningful information within the overall video, thus resulting in lower recall value. As such, the temporal sentence localization methods cannot be directly applied to the video thumbnail generation task.

\begin{table}[!tb]\small
\setlength{\abovecaptionskip}{0.cm}
\setlength{\belowcaptionskip}{-0.cm}
\centering
\caption{ \small Ablation studies on the different components in GTP.}
\begin{tabular}{m{1.6cm} m{0.95cm}<{\centering} m{0.95cm}<{\centering} m{0.95cm}<{\centering} m{0.95cm}<{\centering}}
\hline
Method & Precision & Recall  & F1  & IoU \\
\hline
GTP-G & 0.5053 & 0.5384 & 0.5100 & 0.3756 \\
GTP-P &   0.4071 & 0.4780  & 0.4310  & 0.3043 \\
GTP-C &  0.4968 &  0.4475 &  0.4582 & 0.3237 \\
\bfseries{GTP}  & \textbf{0.5055} & \textbf{0.5742} & \textbf{0.5285} & \textbf{0.3933} \\
\hline
\end{tabular}
\label{table:ablation_study}
\vspace{-0.1in}
\end{table}

\subsection{Analysis of the GTP Model}
\vspace*{0.5mm}
\noindent \textbf{Ablation Studies on the GTP Components.} To verify the contribution of each part of our proposed GTP model, we perform three ablation studies as follows.

(1) \textbf{GTP-G}: We drop the sentence specified video graph convolutional network, and directly feed the concatenation of the average feature of words and video clip feature into the temporal conditioned pointer network.

(2) \textbf{GTP-P}: We drop the temporal conditioned pointer network, and instead establish a 0-1 classifier on the graph convolved video features $\mathbf{H}^G$ to predict the probability of selecting each video clip as the video thumbnail. The top-5 ranked clips with the highest probabilities are concatenated as the final video thumbnail.

(3) \textbf{GTP-C}: We remove the temporal conditioned constraint in the proposed temporal conditioned pointer network. In this case, the selected video clips will further be post-processed by dropping the repetitive ones to produce the final video thumbnail.

Table \ref{table:ablation_study} lists the results
of the aforementioned ablation studies. It can be observed that our full model GTP outperforms all its variants, which clearly verifies the effectiveness of our proposed sentence specified video graph convolutional network and temporal conditioned pointer network. Concretely, the graph convolution establishes sentence specified relationships between different video clips and links the semantically related ones, which thereby supports the in-video reasoning when selecting video clips according to the given sentence semantics.
The temporal conditioned pointer network learns the video thumbnail selection pattern from the training dataset, which can flexibly determine the video clip selection and termination based on the former predictions.
In contrast, GTP-P drops the pointer network and takes the video clip ranking strategy. In this case, the temporal and contextual information within video thumbnails are not fully characterized and the video thumbnail lengths are also fixed to a pre-defined value (5 clips), which inevitably leads into inferior results and makes the video thumbnail generation quite inflexible. Moreover, although the temporal conditioned constraint is simple, it can naturally avoid the disordered and repetitive video clips, and further preserves the logical relations and semantic coherence of  the generated video thumbnails. Therefore, incorporating this constraint from GTP-C to GTP makes a significant performance improvement for the overall model.

\begin{table}[!tb]\small
\setlength{\abovecaptionskip}{0.cm}
\setlength{\belowcaptionskip}{-0.1cm}
\centering
\caption{ \small Ablation studies on the graph convolution layers in GTP.}
\begin{tabular}{m{1.55cm} m{0.95cm}<{\centering} m{0.95cm}<{\centering} m{0.95cm}<{\centering} m{0.95cm}<{\centering}}
\hline
Method & Precision & Recall  & F1  & IoU \\
\hline
GTP-1 & 0.5028 & 0.5686 & 0.5245 & 0.3880 \\
\textbf{GTP-2}  & \textbf{0.5055} & \textbf{0.5742} & \textbf{0.5285} & \textbf{0.3933} \\
GTP-3 &  0.5036 &  0.5710 & 0.5257 & 0.3899 \\
GTP-4 &  0.4985 &  0.5677 & 0.5216 & 0.3854 \\
\hline
\end{tabular}
\label{table:ablation_study_GCNLayer}
\end{table}

\noindent \textbf{Ablation Studies on the Number of Graph Convolution Layers.} Table \ref{table:ablation_study_GCNLayer} lists the results of our proposed GTP model with different numbers of graph convolution layers. It can be observed that GTP with two layers of graph convolutions achieves the best results. When adding more graph convolution layers, the overall performances gradually decrease but still stay stable, with narrow margins compared to the best. The main reason may be that overfitting can become an issue as the number of parameters increases with model depth \cite{Kipf2016Semi}.

\begin{figure*}
\centering
\includegraphics[width=6.0in]{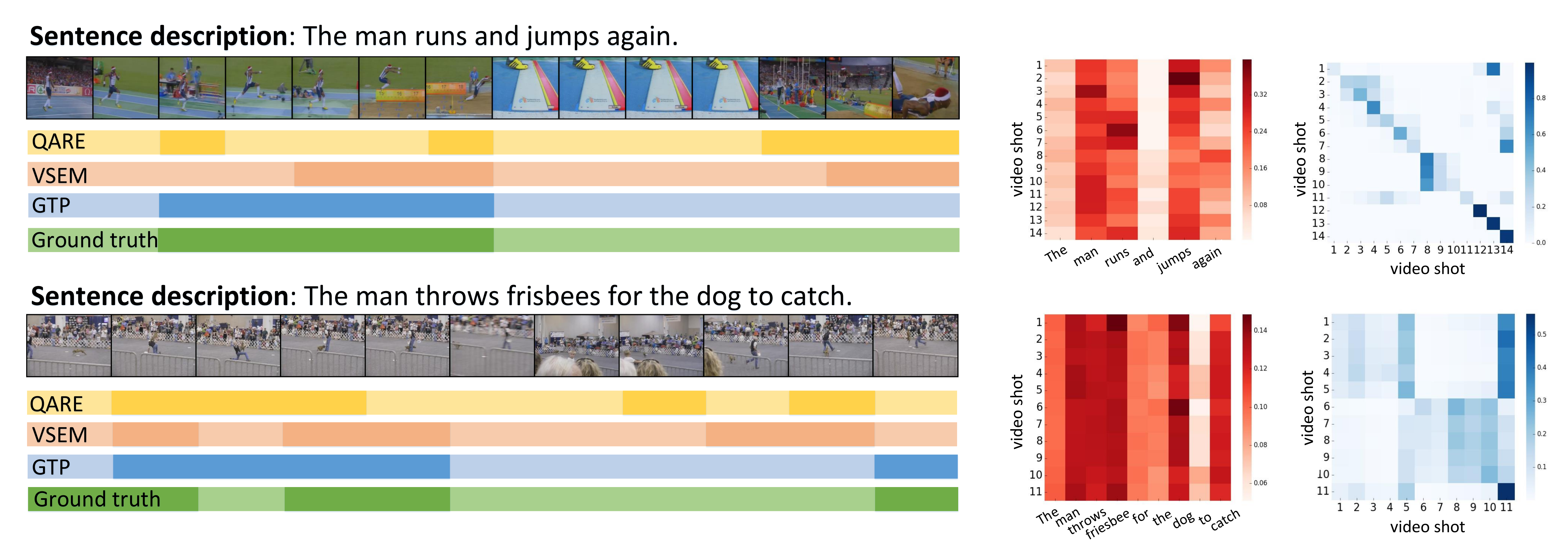}
\caption{ Qualitative examples for sentence specified dynamic video thumbnail generation. On the left, we use different color bars to show the video clip selection results for different methods, with the selected video clips highlighted in darker colors. Ground-truth video thumbnails are indicated by green color. On the right, we provide two kinds of heat maps (red and blue) to illustrate the word-by-clip attention matrix and the video clip adjacency matrix, respectively.}
\label{fig:quality}
\end{figure*}

\begin{figure}
\centering
\includegraphics[width=3.4in]{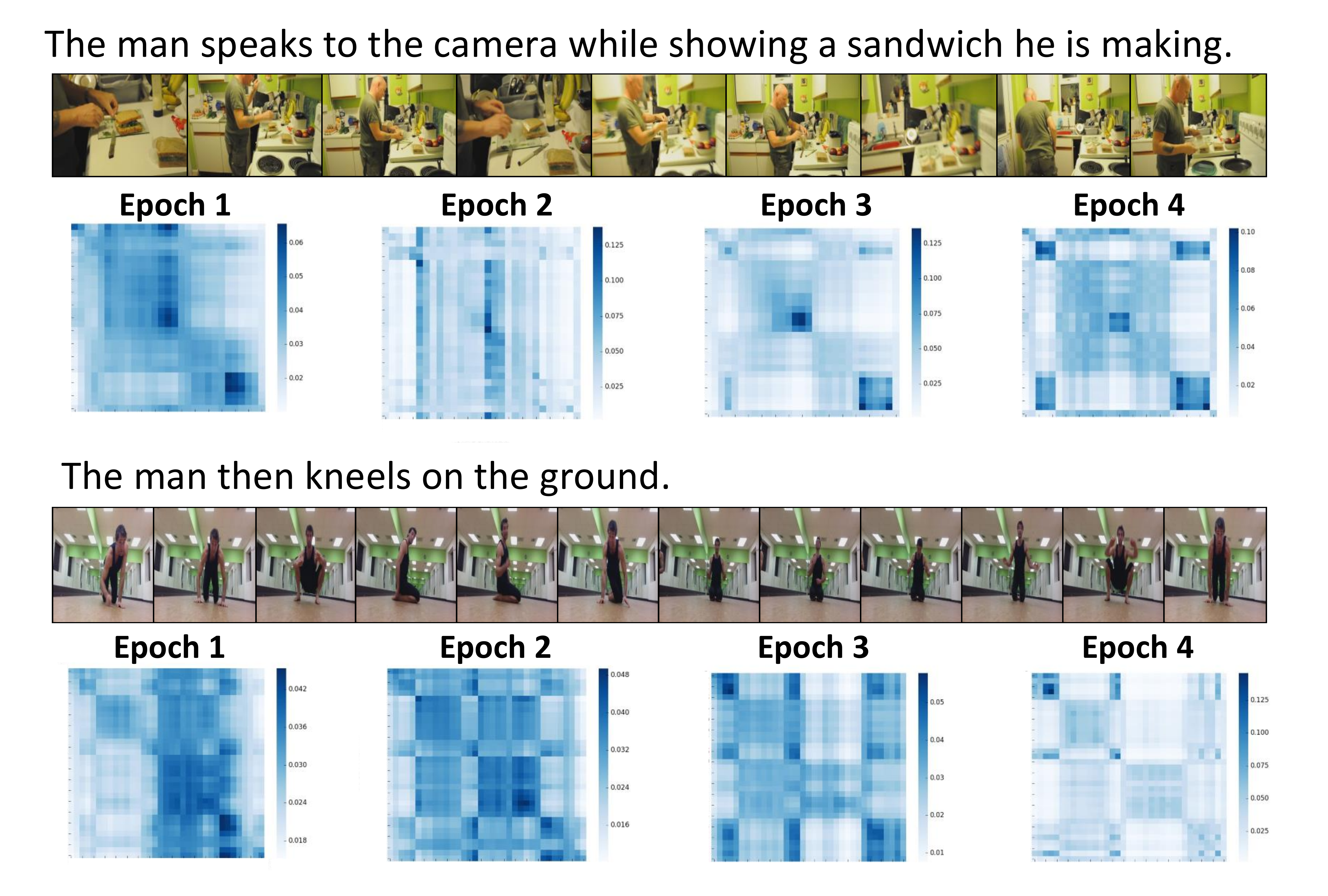}
\caption{ Evolution of the learned adjacency matrices during the sentence specified video graph convolution. The graph edge, representing video clip relationships, are more clearly learned along with the model training procedure. }
\label{fig:video_graph}
\end{figure}

\subsection{Qualitative Results}
\textbf{Video Thumbnail Generation Examples.} Several qualitative examples for sentence specified dynamic video thumbnail generation are shown in Figure~\ref{fig:quality}. It can be observed that the selected video clips of our GTP model are more semantically consistent with the given sentence description. Even in the second example, the ground-truth thumbnails are divided into three separate parts, our GTP can still predict the positions of them accurately. It indicates that our GTP not only measures the semantic correlations between video clips and sentences, but also captures the long range dependencies and internal relationships of videos, and thus can generate video thumbnails providing good content previews of the original videos.

For better demonstrating the word-by-clip interaction and the video graph convolution in the video thumbnail generation procedure, we also provide two kinds of heat maps (red and blue) in Figure \ref{fig:quality} to illustrate the word-by clip attention matrix and the video clip adjacency matrix, respectively. From the word-by-clip attention matrix, it can be observed that some words with higher attention weights well match the video contents. For example, in the first qualitative example, the action ``\texttt{\small man runs and jumps}'' appears in the 3 $\sim$ 7 video clips, and accordingly the concepts  ``\texttt{\small man}'', ``\texttt{\small runs}'' and ``\texttt{\small jumps}'' get higher attention values in these video clips. For the stop words like ``\texttt{\small the}'' and ``\texttt{\small and}'', their attention weights are very small and present an even distribution across the whole video.

For the video clip adjacency matrix, the values in the diagonal region are always higher than the others. It is consistent with the fact that video clips always have higher similarities with their adjacent clips. Additionally, for the second qualitative example,
the last video clip is highly correlated to the first 5 clips under the sentence semantics, illustrating high entry values in the adjacency matrix. Based on the adjacency matrix, our GTP performs reasoning on the video clip graph with graph convolution operation, and thus it can easily link the last video clip to the first 5 video clips. This can also provide an interpretation of why our proposed GTP can accurately predict the position of the separated last video clip.

\noindent \textbf{Video Clip Graph Learning}. To investigate whether our GTP model can learn the sentence specified video clip graph structure in the model training procedure, we select two samples in our training set, and record the evolution of their corresponding video clip adjacency matrices in different training epochs, which are illustrated in Figure~\ref{fig:video_graph}. We can observe that the adjacency matrices tend to an even distribution at Epoch 1.  Along with the model training procedure, the block boundaries gradually show up clearly in the adjacency matrices, which means that the video graph structures are gradually learned. Meanwhile, by examining video contents with respect to the learned adjacency matrices, we can find that video clips linked with higher edge values also present strong semantic correlations. It indicates that our model can indeed learn the sentence specified semantic relationships between different video clips, and further facilitates the video thumbnail generation.

\section{Conclusions}

In this paper, we defined a distinctively new task, namely sentence specified dynamic video thumbnail generation, which aims at selecting and synthesizing several video clips from video to constitute the video thumbnail, such that the video thumbnail semantically corresponds to the given sentence description. To facilitate the proposed video thumbnail generation task, we created a new dataset by re-annotating the videos in the ActivityNet Caption dataset. Furthermore, we proposed a novel GTP model, leveraging the graph convolution operation to explore the sentence specified semantic relationships between different video clips. The informative video thumbnail is thereafter sequentially predicted by a novel temporal conditioned pointer network. Extensive experimental results demonstrate the superiority of our proposed model, which outperforms baseline methods with considerable margins.

\section{Acknowledgments}
This work was supported by National Program on Key Basic Research Project No. 2015CB352300, National Natural Science Foundation of China Major Project No.U1611461 and Shenzhen Nanshan District Ling-Hang Team Grant under No.LHTD20170005.

\bibliographystyle{ACM-Reference-Format}
\balance

\clearpage

\appendix

This supplemental material includes the following contents:
\begin{itemize}
    \item The annotation details of the sentence specified video thumbnail dataset.
    \item Dataset statistical analysis.
    \item More qualitative results of the proposed GTP model.
\end{itemize}

\section{The Dataset Annotation Detail}

\begin{figure}[htbp]
    \centering
    \includegraphics[width=3.2in]{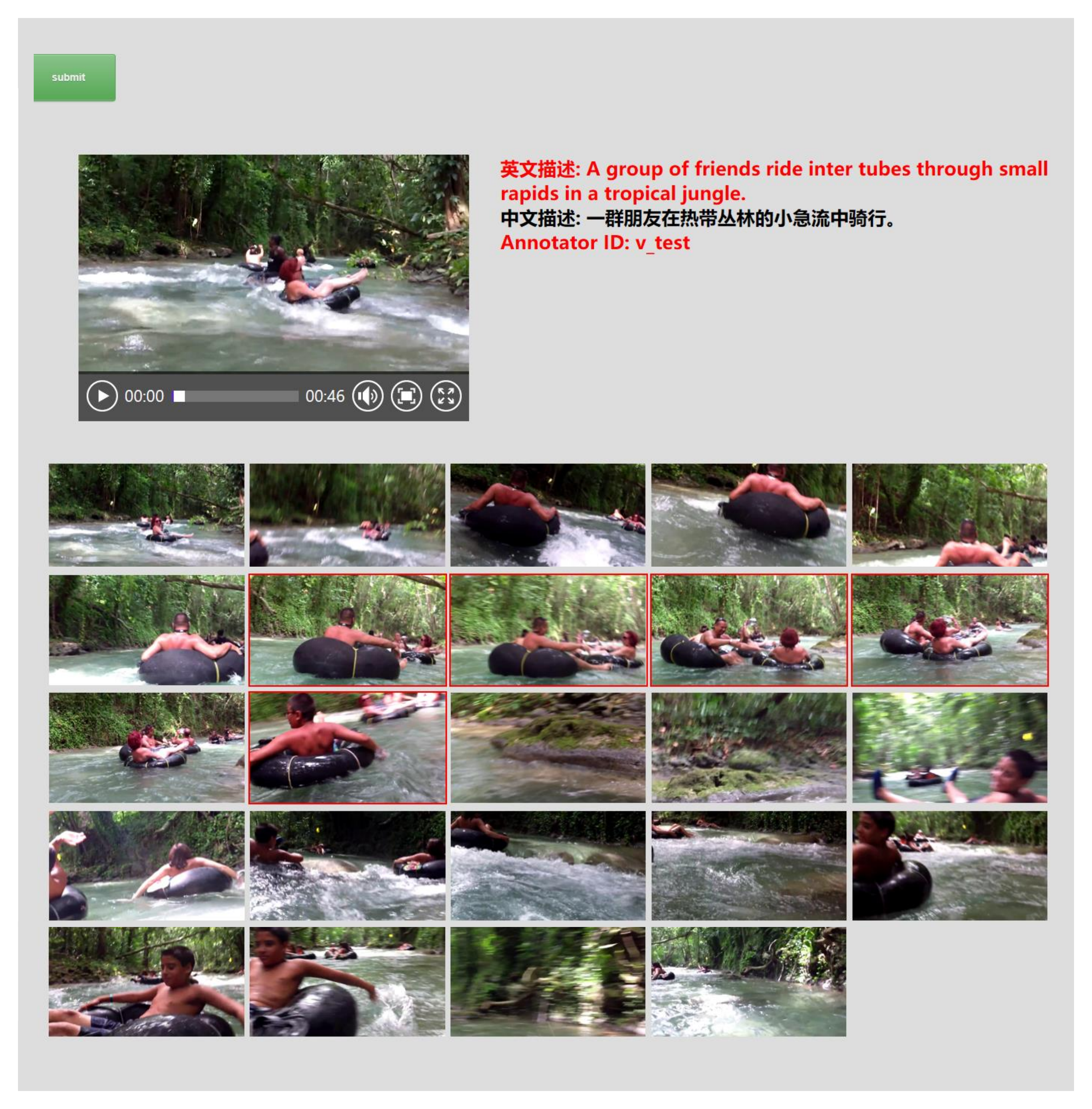}
    \caption{The annotation interface for the sentence specified dynamic video thumbnail generation task.}
    \label{fig:website}
\end{figure}

Figure~\ref{fig:website} illustrates our implemented annotation website for the sentence specified dynamic video thumbnail generation task. For each video and its paired sentence description in our collected dataset, we place them on the website simultaneously for the convenience of the annotation participants' browsing. Moreover, in order to speed up the annotation, we evenly split the video into 2-second video clips (We split the video into 2-second length clips mainly because we find that the smallest video thumbnail gifs in some video websites like YouTube are 1 to 2 seconds long), and all these video clips are displayed in their chronological order. Participants are required to select no more than 5 video clips that semantically correspond to the sentence description to compose the video thumbnail. The video clip will be highlighted in red bounding box after selected. The selected video clips are not required to be consecutive in time. If one participant finishes the video clip selection for the current video-sentence pair, he (or she) only needs to click the ``submit'' button to proceed to the next annotation task.

The annotations of different participants are completely independent, with the video-sentence pairs randomly illustrated on the website. There are 10,204 video-sentence pairs in our collected dataset, and we ensure that each pair will have 4 video thumbnail annotations from 4 different participants. Therefore, we totally get $4\times10,204= 40,816$ annotation results for our constructed dataset.

\begin{figure*}[htbp]
\setlength{\abovecaptionskip}{0.cm}
\setlength{\belowcaptionskip}{-0.cm}
\centering
\includegraphics[width=6.6in]{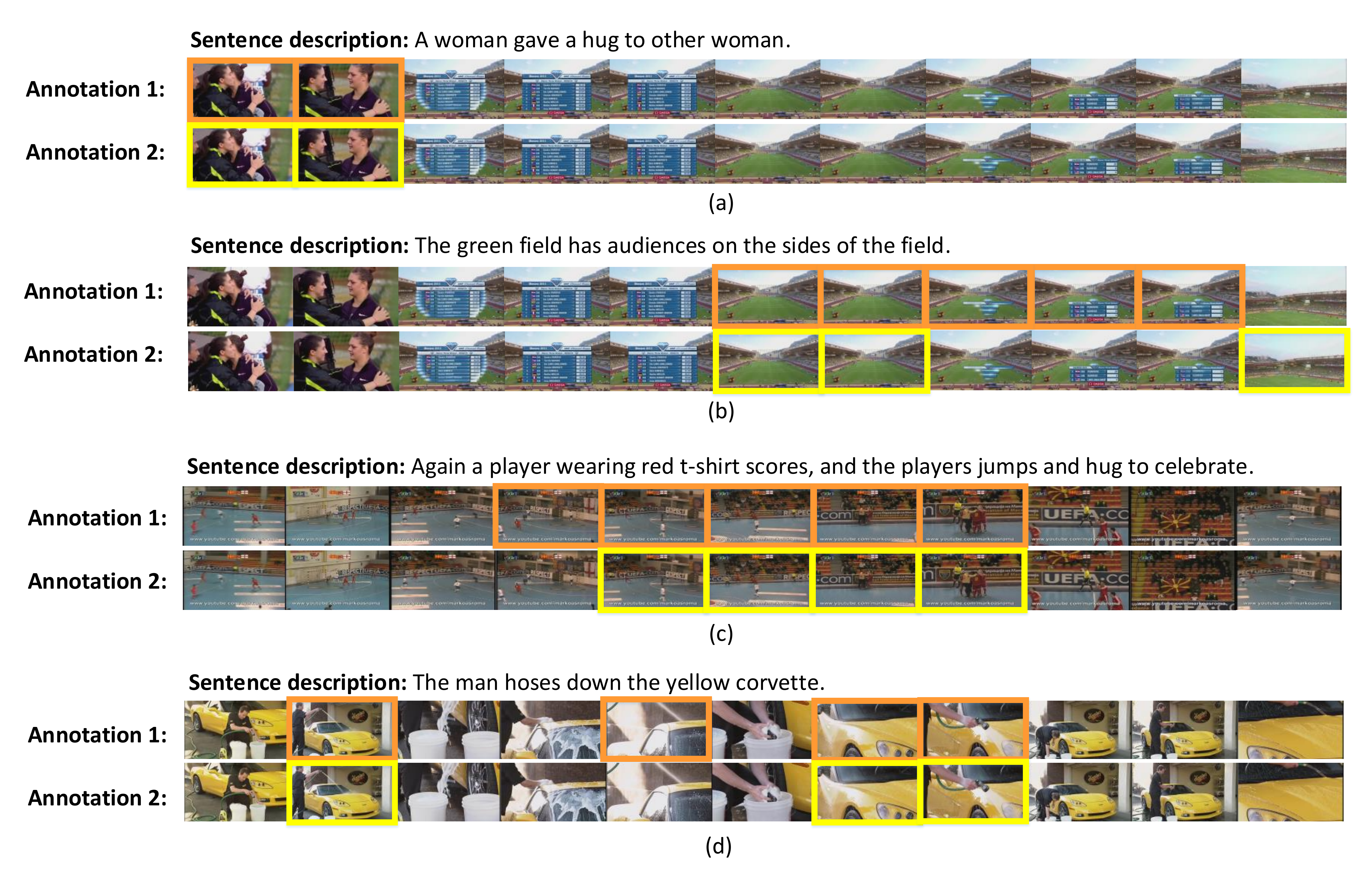}
\caption{ Video thumbnail annotation examples. For each showing video-sentence pair, we provide two video thumbnail annotations, and the selected video clips in these two annotations are highlighted with orange and yellow bounding boxes, respectively. }
\label{fig:dataset}
\end{figure*}

Some video thumbnail annotation examples are shown in Figure \ref{fig:dataset}. For each showing example, we provide two video thumbnail annotations, and the selected video clips in these two annotations are highlighted with orange and yellow bounding boxes, respectively. We can observe that in example (a), the two annotations are exactly the same, while in other examples, the annotations are partially aligned with each others. It illustrates that when annotating video thumbnails, different participants have different opinions, making the differences between the annotated video thumbnails. However, the jointly selected video clips also indicate that the participants still have their common cognition for the given sentence descriptions. In addition, example (a) and example (b) share the same video but are with different sentence descriptions. We can see that the sentence descriptions highly influence the resulting video thumbnails and cause great discrepancy, which further verifies that it is very necessary to generate specific video thumbnails for different sentences.

\begin{figure}
    \centering
    \includegraphics[width=\columnwidth]{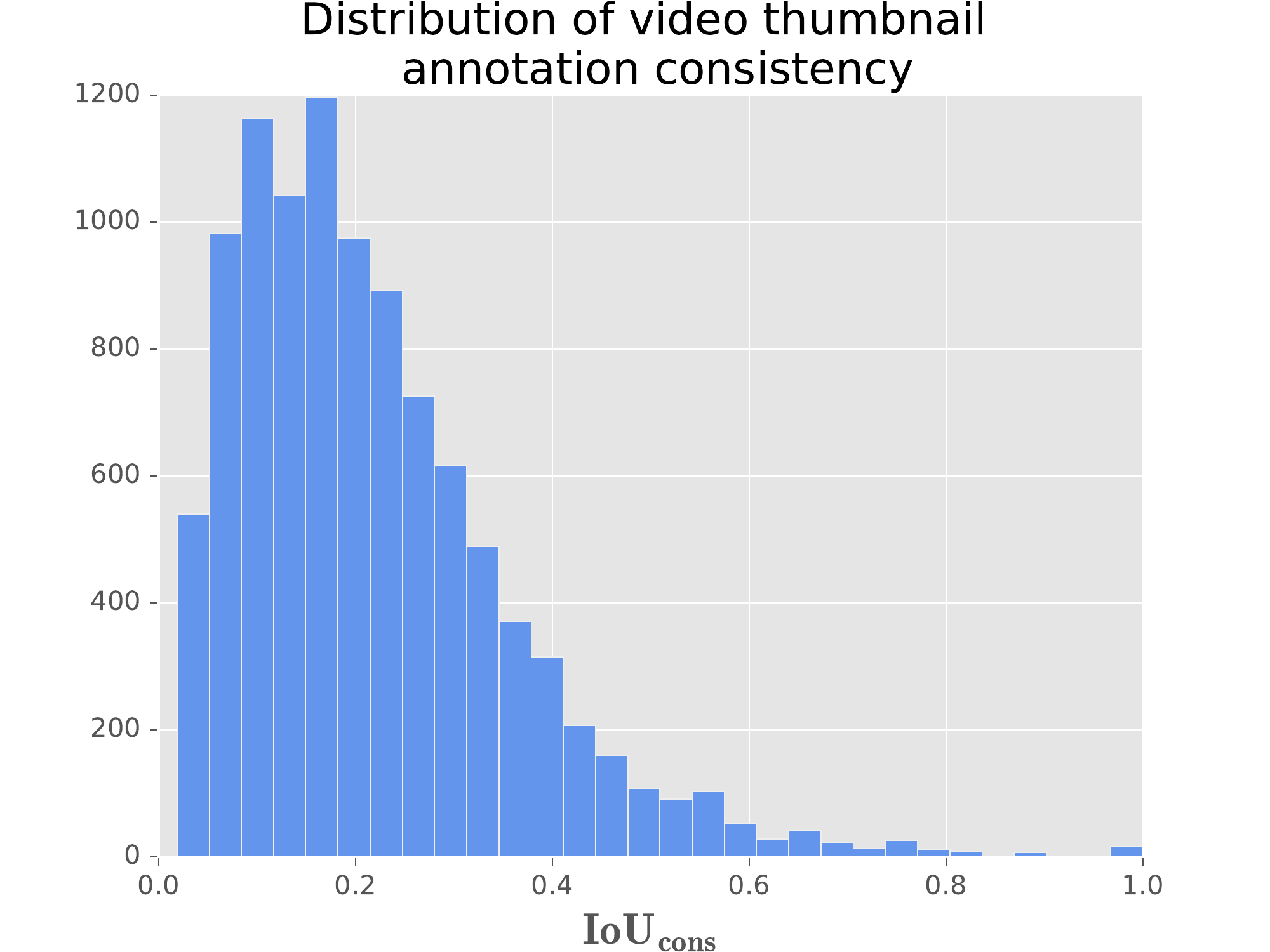}
    \caption{The video thumbnail annotation consistency distribution over all the video-sentence pairs.}
    \label{fig:consistency}
\end{figure}

\vspace*{2mm}
\section{Dataset Statistical Analysis}

\textbf{Video Length.} The minimal, maximal, and average video lengths over all the videos in our constructed dataset are 20.0s, 238.4s and 60.7s, respectively. The average length of the annotated video thumbnails is 8.7s.


\noindent \textbf{Video Thumbnail Annotation Consistency.} As indicated in Figure \ref{fig:dataset}, video thumbnail annotation is a very subjective task, with different annotation participants having different opinions. To measure the consistency of the selected video thumbnails between different participants, we define a metric $IoU_{cons}$ as follows:
\begin{equation}
\small{
\begin{split}
    IoU_{cons}(k,i) = \frac{1}{3} & \sum_{j \not= i, j = 1}^4 \frac{ \| Intersection(A_i^k,A_j^k) \|}{ \| Union(A_i^k,A_j^k) \|} \\
    IoU_{cons}(k) & = \frac{1}{4} \sum_{i=1}^4 IoU_{cons}(k,i)
\end{split}
}
\end{equation}
Here $A_i^k$ means the set of selected video clips composing the $i$-th annotated video thumbnail for the $k$-th video-sentence pair. $IoU_{cons}(k,i)$ indicates the annotation consistency between the $i$-th annotated video thumbnail and all the other annotations for the $k$-th video-sentence pair. $IoU_{cos}(k)$ means the average annotation consistency of the 4 video thumbnail annotations for the $k$-th video-sentence pair. If the selected video clips of all the annotations are exactly the same, the value of $IoU_{cos}(k)$ will be equal to 1. The annotation consistency distributed over all the video-sentence pairs is illustrated in Figure~\ref{fig:consistency}. It can be observed that for most of the video-sentence pairs, the selected video clips of different participants do not have a exact match, but there are still some clips that are jointly selected by several participants. It further demonstrates that the video thumbnail generation is an indeed subjective task, while people still express their consensus to generate the thumbnail with respect to the given sentence descriptions.

\noindent \textbf{Ground Truth.} Since there are 4 video thumbnail annotations for each video-sentence pair, we take the annotation result with the highest consistency $IoU_{cons}(k,i)$ among the 4 annotations as the ground truth during the training process. While in the testing stage, the predicted video thumbnail will be evaluated with respect to all the 4 annotations.


\vspace*{3mm}
\section{Qualitative Results}

\begin{figure*}[htbp]
    \centering
    \includegraphics[width=6.0in]{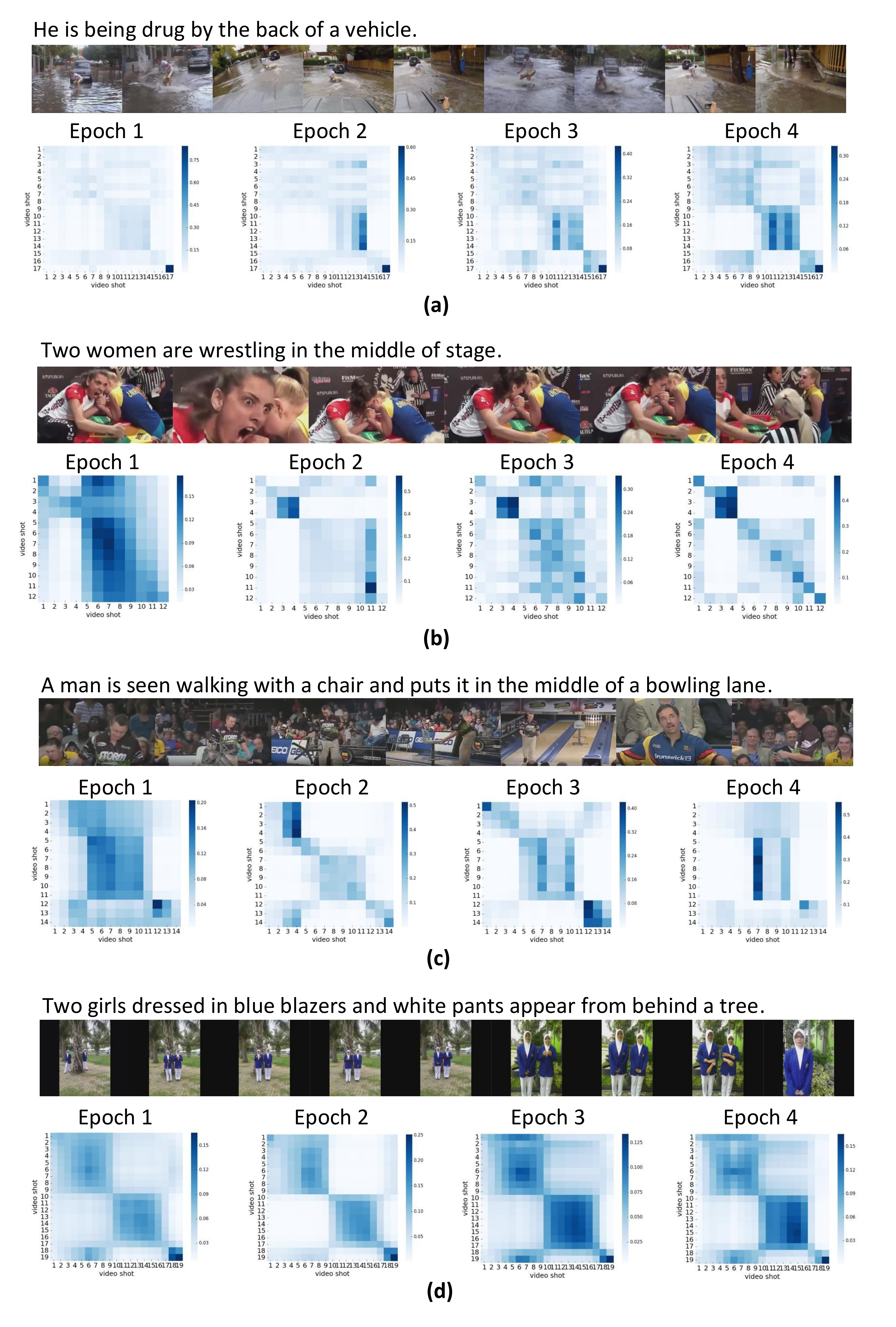}
    \caption{Evolution of the learned video clip adjacency matrices during the sentence specified video graph convolution.}
    \label{fig:video_graph_evolution}
\end{figure*}

\noindent \textbf{Evolution of the Sentence Specified Video Clip Graph.}
Figure \ref{fig:video_graph_evolution} shows the evolution of 4 groups of video clip adjacency matrices in our GTP model training procedure. We can observe that the first two qualitative examples (a) and (b) present similar evolution process with the examples we have shown in the main paper. The adjacency matrices tend to a even distribution
at the initial model training stage, and along with the model training procedure the block boundaries gradually show up clearly. In contrast, in the qualitative examples (c) and (d), the sentence specified video clip graph structures have been initially learned in Epoch 1, with the following training epochs only adjusting and emphasizing the learned video clip relationships. Overall, all of the above results verify that our GTP model can indeed learn the sentence specified video clip graph according to the sentence and video semantics.

\begin{figure*}[htbp]
    \centering
    \includegraphics[width=6.3in]{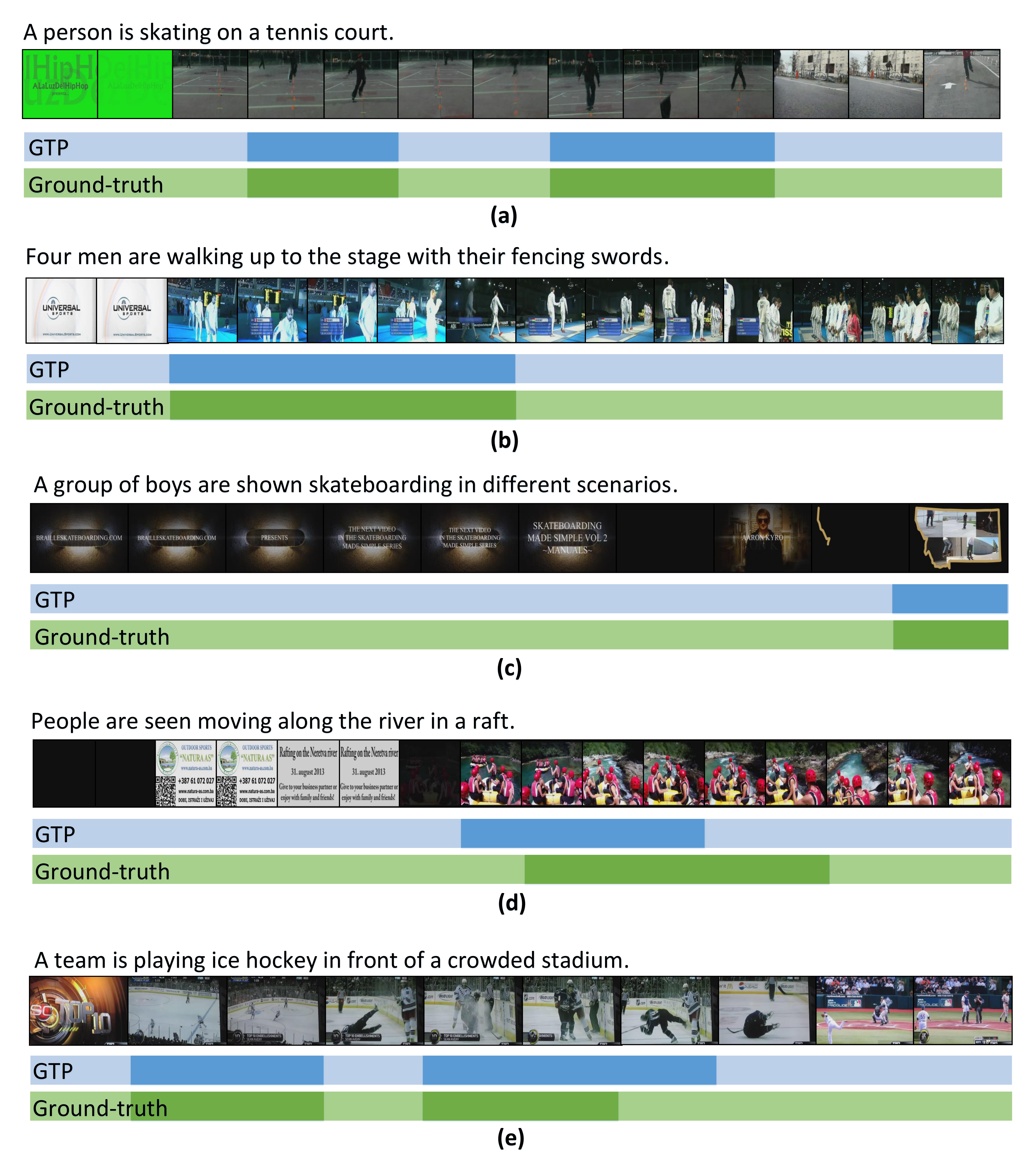}
    \caption{Qualitative results of our proposed GTP model for sentence specified dynamic video thumbnail generation. Blue bars show the video thumbnail generation results for our proposed GTP model, with the selected video clips highlighted in darker colors. Green bars show the ground-truth video thumbnails.}
    \label{fig:thumbnail_qualitative}
\end{figure*}

\noindent \textbf{Video Thumbnail Generation Results of the GTP Model.}
Figure~\ref{fig:thumbnail_qualitative} illustrates some qualitative results of our proposed GTP model for the sentence specified dynamic video thumbnail generation. We can observe that the selected video clips by GTP are consistent with the clips in the ground-truths, which indicates the effectiveness of our proposed GTP model. Meanwhile, the generated video thumbnails are quite flexible. As shown in case (a) and (e), the video thumbnails are temporally inconsecutive and provide a good preview of the overall video content. Comparing the show case (c) to others, we can find that the lengths of video thumbnails are also not fixed. Since most video contents shown in case (c) are irrelevant to ``\texttt{skateboarding}'' described by the sentence, GTP only selects the last clip that presents the matching activity. 

Besides, the predicted video thumbnail in case (d) does not exactly match the ground-truth annotation. The main reason lies on the indistinguishable video scenes in the video. From the 8-th video clip in case (d) to the end of the video, all the middle clips present the same scene of ``\texttt{people rafting}''. Therefore, not only the GTP model, the annotators are also hard to decide which clip to choose. However, since all these clips are matched with the sentence description, the generated video thumbnail by our proposed GTP is still reasonable and accurate.

\end{document}